\documentclass{article}

\usepackage[preprint]{neurips_2023}

\usepackage[utf8]{inputenc} 
\usepackage[T1]{fontenc}    
\usepackage{hyperref}       
\usepackage{url}            
\usepackage{booktabs}       
\usepackage{amsfonts}       
\usepackage{nicefrac}       
\usepackage{microtype}      
\usepackage{xcolor}         

\usepackage{amsmath}
\usepackage{wrapfig, lipsum, booktabs}
\usepackage{graphicx}
\usepackage{natbib}
\usepackage{caption}
\title{Reformulation of RBM to Unify Linear and Nonlinear Dimensionality Reduction}

\author{%
  Jiangsheng You\\
  Aspen Technology Inc\\
  \texttt{jason.you@aspentech.com, jshyou@gmail.com}
  \And
  Chun-Yen Liu \\
  Aspen Technology Inc \\
  \texttt{mark.liu@aspentech.com}
}

\begin{document}

\maketitle

\begin{abstract}
A restricted Boltzmann machine (RBM) is a two-layer neural network with shared weights and has been extensively studied for dimensionality reduction, data representation and recommendation systems in the literature. The traditional RBM requires a probabilistic interpretation of the values on both layers and a Markov chain Monte Carlo (MCMC) procedure to generate samples during the training. The contrastive divergence (CD) is efficient to train the RBM but its convergence has not been proved mathematically. In this paper, we investigate the RBM by using a maximum a posteriori (MAP) estimate and the expectation–maximization (EM) algorithm. We show that the CD algorithm without MCMC is convergent for the conditional likelihood object function. Another key contribution in this paper is the reformulation of the RBM into a deterministic model. Within the reformulated RBM, the CD algorithm without MCMC approximates the gradient descent (GD) method. This reformulated RBM can take the continuous scalar and vector variables on the nodes with flexibility in choosing the activation functions. Numerical experiments show its capability in both linear and nonlinear dimensionality reduction, and, for the nonlinear dimensionality reduction, the reformulated RBM can outperform principal component analysis (PCA) by choosing the proper activation functions. Finally, we demonstrate its application to vector-valued nodes for the CIFAR-10 dataset (color images) and the multivariate sequence data, which cannot be configured naturally with the traditional RBM. This work not only provides theoretical insights regarding the traditional RBM but also unifies the linear and nonlinear dimensionality reduction for scalar and vector variables.
\end{abstract}

\section{Introduction} \label{introduction}
The restricted Boltzmann machine (RBM) was introduced in 1986 under the name of \textit{Harmonium} in these early works [\cite{Hinton84, Ackley85, Hinton86, Smolensky86}]. Hinton and his coworkers proposed the contrastive divergence (CD) method to train the RBM [\cite{Hinton02, Hinton06}]. CD has been an efficient and successful algorithm to train the RBM numerically. In fact, to the best of our knowledge, the paper by Hinton and Salakhutdinov [\cite{Hinton06}] was the first publication to report better results with  the neural network-based algorithm (1.2\% error rate) compared with the support vector machine (SVM) (1.4\% error rate) for MNIST dataset. This achievement increased interests in the application of deep learning to recommendation systems [\cite{Ruslan07}], paved the path for more rapid progress in the research and application of deep network architecture, and eventually led to the artificial intelligence boom by [\cite{Alex17}] and subsequent breakthroughs from extensive use of GPU and new network architecture such as Transformer [\cite{Vaswani17}]. For a more comprehensive introduction of RBMs, we refer the readers to other review articles and textbooks [\cite{Fischer14, Goodfellow15, Harshvardhan20, BondTaylor22}]. 
\begin{figure}[ht]
\hfill
\begin{center}
\includegraphics[width=2.5in]{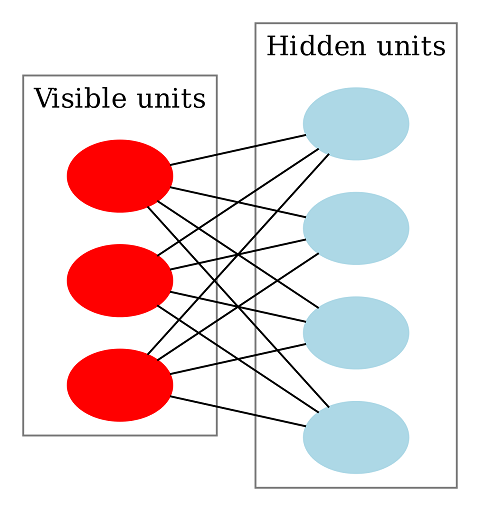}
\end{center}
\vskip -0.1in
\caption{RBM layers}
\label{rbm_layers}
\end{figure}

\noindent
The RBM is a two-layer network model with no connections between nodes in the same layer, as shown in Figure \ref{rbm_layers}. In the original RBM, each node associates with a binary random variable following the Gibbs distribution jointly. In this section, we cite the detailed mathematical exposition on the probabilistic model for the RBM from [\cite{Fischer14}]. Let ${\bf v}$ and ${\bf h}$ represent the random variables in the visible and hidden layers, respectively. For biases ${\bf a}$ and ${\bf b}$, and a weight matrix ${\bf W} = \{w_{ji}\}$, the Gibbs distribution is described by the following exponential function
\begin{equation}
\label{eq_gibbs}
{p}(\bf{v, h};\,\bf{a, b, W})=\frac{1}{Z}\exp(\bf{a}^\mathrm{T}\bf{v}+\bf{b}^\mathrm{T}\bf{h}+\bf{h}^\mathrm{T}\bf{Wv})
\end{equation}
\noindent
where $\bf{Z}$ is the partition function for random variables $\bf{v}$ and $\bf{h}$, $\bf{a}$ and $\bf{b}$ are the vectors with dimensions of $m$ and $n$, respectively, and $\bf{W}$ has dimensions of $n\times m$. To shorten the notations, we let $\boldsymbol{\theta}=(\bf{a, b, W})$ represent all the parameters and $S=\{v\}$. Each $v$ represents an observation of $\bf{v}$ and $K$ is the total number of observations. The unconditional likelihood function ${L\bf (\boldsymbol{\theta};\, v)}$ can be defined as
\begin{equation}
\label{eq_likelihood_eqn}
{L\bf (\boldsymbol{\theta};\, v)} = {\mathop{\prod }_{v \in S}\left[{\mathop{\mathrm{\sum }}_{\bf h}} p(\bf{v, h};\, \boldsymbol{\theta})\right]} 
\end{equation}

\noindent
Based on previous works [\cite{Hinton02, Hinton10, Fischer14}], finding the solution for the RBM is equivalent to solving the maximum likelihood problem $\mathop{\mathrm{argmax}}_{\boldsymbol{\theta}}  {L(\boldsymbol{\theta};\, \bf{v})}$, and the average derivatives of the log-likelihood over the training set $S$ are
\begin{equation}
\label{eq_likelihood_grad}
\frac{1}{K} \sum_{v\in S}\frac{\partial \ln{L\bf (\boldsymbol{\theta};\, v)}}{\partial w_{ji}} \propto \left[{<v_ih_j>}_{data} - {<v_ih_j>}_{model}\right]
\end{equation}
\noindent
When the RBM is interpreted as a stochastic neural network with the sigmoid function, $\mathrm{sigm}(\cdot)$, as the activation function, the conditional probability of $\bf{v}$ and $\bf{h}$ are obtained as follows 
\begin{equation}
\label{eq_cond_prob}
\begin{aligned}
&p(h_j=1 | {\bf v}) = \mathrm{sigm}(\sum_i^mw_{ji}v_i+b_j) \\
&p(v_i=1 | {\bf h}) = \mathrm{sigm}(\sum_j^nw_{ji}h_j+a_i)
\end{aligned}
\end{equation}
\noindent
We refer readers to the previous works [equations 8-10 of \cite{Fischer14}] for the comprehensive derivations of \eqref{eq_likelihood_grad} and \eqref{eq_cond_prob}, which will not be reproduced here. We note that the assumptions behind \eqref{eq_cond_prob} are the Gibbs distribution and binary values of $\bf{v}$ and $\bf{h}$. Hence, it is unclear if such factorized probabilities in \eqref{eq_cond_prob} can be derived for other types of distributions. Here, we list three assumptions of the traditional RBM below:
\begin{itemize}
  \item The partition function ${\bf Z}$ is calculated over ${\bf v}$ and ${\bf h}$ independently, and the distributions of ${\bf v}$ and ${\bf h}$ depend on the unknown model ${\boldsymbol \theta}$.
  \item Training data $S$ are the observations of ${\bf v}$ and the true distribution of ${\bf v}$ is unknown. Thus, only the empirical distribution of ${\bf v}$ can be computed.
  \item From the definition of Markov random fields (MRF), $\bf{v}$ and $\bf{h}$ should be independent but the derivation of equations \eqref{eq_likelihood_grad} and \eqref{eq_cond_prob} needs a conditional probability from latent variables as described in Section 2 of [\cite{Fischer14}].
\end{itemize}
\noindent
The so-called CD-$n$ learning algorithm with $n$-cycles of Markov chain Monte Carlo (MCMC) was proposed to approximate ${<v_ih_j>}_{model}$ in equation \eqref{eq_likelihood_grad} to find the solution for the RBM [\cite{Hinton02, Fischer14, Jacob20}]. Numerically, the researchers found that CD-$1$ is sufficient to find the optimal solution for the RBM [\cite{Hinton06, Salakhutdinov09, Hinton10, Yamashita14, Taniguchi23}]. However, theoretically, infinite cycles of MCMC, i.e., CD-$\infty$, are required to accurately estimate ${<v_ih_j>}_{model}$. Though the CD learning algorithm is numerically effective and convergent [\cite{Hinton02, Hinton06, Salakhutdinov09, Hinton10}], the gap between numerical implementations and the theoretical understanding was not fulfilled. Untill now, little has been known from a theoretical viewpoint on the convergence of CD-$1$ in terms of increasing ${L\bf (\boldsymbol{\theta};\, v)}$, except that CD can not be the gradient of any function and may not converge under certain conditions [\cite{Carreira05, Bengio09, Sutskever10, Fischer10, Fischer11}].

In this paper, we revisit the probabilistic RBM model and apply the expectation maximization (EM) algorithm for a maximum a posteriori (MAP) estimate to study the RBM model. In Section \ref{probabilistic_rbm}, we intend to understand the mathematics behind CD is and explain why CD-$1$ is sufficient to obtain the solution of the traditional RBM. In Section \ref{deterministic_rbm}, we reformulate the RBM as a deterministic model, which offers several benefits, such as eliminating the limitations of the binary input variables and the sigmoid activation function. Finally, in Sections \ref{image_analysis} and \ref{multivariate_analysis}, we demonstrated the capability of the reformulated RBM in dimensionality reduction and multivariate analysis.

\section{EM Algorithm for the Probabilistic Model} \label{probabilistic_rbm}
\noindent
In contrast to the existing assumption that ${\bf v}$ and ${\bf h}$ are independent random variables, in this paper we regard ${\bf h}$ as latent variables and ${\boldsymbol \theta}$ as parameters to be estimated without using the MRF to define the RBM model. In practice, only limited amount of observations of ${\bf v}$ are known and the conditional distribution of ${\bf h}$ can be calculated by \eqref{eq_cond_prob}. For a given training set $S=\{{v}\}$ and an initial estimate $\boldsymbol \theta^{(0)}$ of $\boldsymbol \theta$ in the probabilistic RBM model, ${\bf h}$ follows the conditional probability $p(\cdot| {S}, {\boldsymbol \theta}^{(0)})$ in Equation \eqref{eq_cond_prob}, and the conditional Gibbs distribution becomes 
\begin{equation}
\label{eq_gibbs_cond}
p_{[{\bf h}\sim p(\cdot| S, {\boldsymbol \theta}^{(0)})]}(\bf{v, h};\,{\boldsymbol \theta})=\frac{1}{Z}\exp(\bf{a}^\mathrm{T}\bf{v}+\bf{b}^\mathrm{T}\bf{h}+\bf{h}^\mathrm{T}\bf{Wv})
\end{equation}
\noindent
We note that the partition function ${\bf Z}$ is calculated under the conditional distribution $p(\cdot| {S}; {\boldsymbol \theta}^{(0)})$ for ${\bf h}$. Following the configuration of the expectation–maximization (EM) algorithm, we consider the expected value of a conditional log-likelihood
\begin{equation}
\label{eq_loglikeli_cond}
Q(\boldsymbol \theta \, |\, \boldsymbol \theta^{(0)}) = \frac{1}{K}\sum_{v\in S}{\bf E}_{[{\bf h}\sim p(\cdot| S, {\boldsymbol \theta}^{(0)})]}\left[\log(p({\bf v, h};{\boldsymbol \theta}))\right]
\end{equation}
\noindent
Based on the derivation of [equation-29 of \cite{Fischer14}], we obtain
\begin{equation}
\label{eq_loglikeli_grad}
\frac{\partial Q(\boldsymbol \theta \,|\, \boldsymbol \theta^{(0)})}{\partial w_{ji}} \propto \left[{<v_ih_j>}_{data} - {<v_ih_j>}_{\boldsymbol \theta^{(0)}}\right]
\end{equation}
Equation \eqref{eq_loglikeli_grad} can then be used to find the maximum parameter $\boldsymbol \theta^{(1)}$ of $Q(\boldsymbol \theta \,|\, \boldsymbol \theta^{(0)})$
\begin{equation}
\label{eq_loglikeli_max}
\boldsymbol \theta^{(1)} = \mathop{\mathrm{argmax}}_{\boldsymbol{\theta}} {Q(\boldsymbol \theta \,|\, \boldsymbol \theta^{(0)})}
\end{equation}
\noindent
Hereafter we use CD-0 to represent the CD without MCMC sampling, which is defined as Algorithm 1 in [\cite{Fischer14}]. According to the convergence theorem of the EM algorithm [\cite{Dempster77, WuCF83, Meng97}], we have the following inequality, indicating the convergence of the CD-0 algorithm in terms of increasing ${L\bf (\boldsymbol{\theta};\, v)}$
\begin{equation}
\label{eq_loglikeli_converge}
\log L(\boldsymbol \theta^{(1)};\, {\bf v}) - \log L(\boldsymbol \theta^{(0)};\, {\bf v}) \ge {Q(\boldsymbol \theta^{(1)} \,|\, \boldsymbol \theta^{(0)})} - {Q(\boldsymbol \theta^{(0)} \,|\, \boldsymbol \theta^{(0)})}
\end{equation}

\noindent
Replacing ${\boldsymbol \theta^{(0)}}$ by ${\boldsymbol \theta^{(1)}}$ and keep improving the estimate of ${\boldsymbol \theta^{(1)}}$ in \eqref{eq_loglikeli_max}, this iteration continuously increases the expectation of the conditional log-likelihood defined by \eqref{eq_loglikeli_cond}, leading to the convergence of ${L\bf (\boldsymbol{\theta^{(1)}};\, v)}$. Summarizing the preceding arguments, we claim:
\begin{itemize}
  \item For an initial $\boldsymbol \theta^{(0)}$, CD-$0$ provides an estimate of $\boldsymbol \theta^{(1)}$ of solution \eqref{eq_loglikeli_max}.
  \item CD-$0$ increases ${L\bf (\boldsymbol{\theta};\, v})$ according to the EM convergence theorem \eqref{eq_loglikeli_converge}.
\end{itemize}

\noindent
{\bf Remark 1.} \emph{Please keep in mind that two critical assumptions in Sections \ref{introduction} and \ref{probabilistic_rbm} are: (1) ${\bf v, h}$ take binary values of $\{0, 1\}$, and (2) they jointly follow the Gibbs distribution. Under these two assumptions, the CD-0 learning can be regarded as the GD method to implement the EM algorithm. In our numerical experiments, the conditional probabilities are used in the right side of \eqref{eq_loglikeli_grad} without MCMC, see previous literature [\cite{Woodford06, BengioY09, Swersky10, Fischer14}] for more discussions about whether using MCMC sampling or not. We note that the iteration discussed in some previous works [\cite{Hinton02, Hinton10, Fischer14}] follows the EM algorithm steps though it was never explicitly elaborated with the MAP framework.}

We refer the readers to previous works for the convergence of EM algorithm [\cite{Dempster77, WuCF83, Meng97}] and for the image reconstruction [\cite{Shepp82, Hart87, You07}]. Compared with two previous works [\cite{Song16, WuM19}] on using the EM algorithm for the RBM, this work does not need the MCMC and a penalty.

\section{Gradient Descent for the Deterministic Model} \label{deterministic_rbm}

In this section, we reformulate RBM as a pair of reciprocal mappings between visible and hidden layers followed by applying the activation functions. Let $R^{m}$, $R^{n}$, $R^{n\times m}$ denote Euclidean spaces with dimensions of $m$ , $n$ , and $n\times m$, correspondingly. Vectors ${X}\in R^{m}$ and ${Y}\in R^{n}$ represent the data on visible and hidden layers, respectively. Matrix $\textbf{W}\in R^{n\times m}$ represents a linear transformation from $R^{m}$ to $R^{n}$. Let ${A_{v}}(\cdot )$ and $ {A_{v}^{'}}(\cdot )$ denote the activation functions and their derivatives in the visible layer, and ${A_{h}}(\cdot ) $ and ${A_{h}^{'}}(\cdot )$ denote the activation functions and their derivatives in the hidden layer. The reciprocal mappings between visible and hidden layers are defined by 
\begin{equation}
\label{eq_mapping_yx}
{Y}={A_{h}}(\textbf{W}{X}+{B_{h}}); \quad {X}={A_{v}}(\textbf{W}^{T}{Y}+{B_{v}})
\end{equation}
\noindent
Here, $Y$ stands for the mapping from $R^{m}$ to $R^{n}$ with the bias of ${B_h}$ and the activation functions of ${A_h}(\cdot )$, and ${X}$ stands for the mapping from $R^{n}$ to $R^{m}$ with the bias of ${B_v}$ and the activation functions of ${A_{v}}(\cdot )$. We note that the values in $Y$ and $X$ can be continuous variables and the activation functions ${A_h}(\cdot )$ and ${A_v}(\cdot )$ do not require any probabilistic interpretations. We use $S=\{v\}$ for the training data, and let $\Theta=\{\textbf{W}, B_h, B_v\}$ be the parameter to be estimated and $E(X;\, \textbf{W}, {B_{h}}, {B_{v}})$ denote the reconstruction error for a single observation $X\in S$, i.e.,
\begin{equation}
\label{eq_se}
\begin{aligned}
E(X;\, \textbf{W}, {B_{h}}, {B_{v}}) = 
\frac{1}{2}\Vert 
{A_{v}}(\textbf{W}^{T}{A_{h}}(\textbf{W}{X}+{B_{h}})+{B_{v}})-{X}\Vert ^{2}
\end{aligned}
\end{equation}
Thus, finding an optimal solution over all samples in $S$ is equivalent to solve the following minimization problem over the training data $S$:
\begin{equation}
\label{eq_se_min}
\mathop{\mathrm{argmin}}_{\textbf{W}, {B_{h}}, 
{B_{v}} }{\mathop{\mathrm{\sum }}_{X\in S}}E(X;\, \textbf{W}, {B_{h}}, 
{B_{v}})
\end{equation}
Model \eqref{eq_se_min} becomes a deterministic model to minimize the reconstruction error over all observations of $S$. More investigations of minimizing the reconstruction error can be found in [\cite{Vincent10}]. We note that, under the concept of recirculation discussed in [\cite{Hinton87}], the reconstruction error is defined between Equation \eqref{eq_mapping_yx} and the original data after the activation by the following equation
\begin{equation}
\label{eq_se_modified}
\begin{aligned}
C(X;\, \textbf{W}, {B_{h}}, {B_{v}})=
\frac{1}{2}\Vert 
{A_{v}}(\textbf{W}^{T}{A_{h}}(\textbf{W}{X}+{B_{h}})+{B_{v}})-{A_{v}}({X})\Vert 
^{2}
\end{aligned}
\end{equation}
The formulation in \eqref{eq_se_modified} will calculate the squared error after applying the activation function ${A_{v}}$ to both original data and the reconstructed data. This configuration may be more plausible from a mathematical sense, but it has not been adopted practically.

\noindent
{\bf Remark 2.} \emph{Model \eqref{eq_se} looks like a least square regression, which is actually not the case. In general, the least square regression is a quadratic function of parameters. However, \eqref{eq_se} is a quartic function with respect to ${\bf W}$, which may lead to more complexity in finding the solution of model \eqref{eq_se_min} due to the nonlinearity of its gradients.}

Function $E(X;\, \textbf{W}, {B_{h}}, {B_{v}})$ in Equation \eqref{eq_se} has a quartic expression and we have not seen an explicit formula of the gradients of \eqref{eq_se} in the literature. In this paper we obtain an explicit formula and have discovered that CD-0 is an approximation of the gradient descent method. For convenience purposes, we introduce three definitions
\begin{equation}
\label{eq_map_def1}
\widetilde{Y}=\textbf{W}{X}+{B_{h}}, \quad \widetilde{X}=\textbf{W}^{T}{A_{h}}(\textbf{W}{X}+{B_{h}})+{B_{v}}, \quad {\Delta X}={A_{v}}({\widetilde{X}})-{X}
\end{equation}
We use $\odot$ to denote the element-wise multiplication between two vectors, and ${Y}{{X}}^T$ is understood as the matrix multiplication between $R^{n\times 1}$ and $R^{1\times m}$. By linear algebra and multivariable calculus, we obtain the gradients of Equation \eqref{eq_se} as follows:
\begin{equation}
\label{eq_se_grad_v}
\frac{\partial }{\partial 
{B_{v}}}E(X;\, \textbf{W}, {B_{h}}, 
{B_{v}})={A_{v}^{'}}(\widetilde{X})\odot {\Delta X}
\end{equation}
\begin{equation}
\label{eq_se_grad_h}
\begin{aligned}
\frac{\partial }{\partial {B_{h}}}E(X;\, \textbf{W}, {B_{h}}, {B_{v}})= 
{A_{h}^{'}}(\widetilde{Y})\odot [\textbf{W}({A_{v}^{'}}(\widetilde{X})\odot {\Delta X})]
\end{aligned}
\end{equation}
\begin{equation}
\label{eq_se_grad_W}
\begin{aligned}
\frac{\partial }{\partial \textbf{W}}E(X;\, \textbf{W}, {B_{h}}, {B_{v}}) = 
{Y}[{A_{v}^{'}}(\widetilde{X})\odot {\Delta X}]^{T} 
+{{A_{h}^{'}}(\widetilde{Y})\odot [\textbf{W}({A_{v}^{'}}(\widetilde{X})\odot {\Delta 
X})]}{X}^{T}
\end{aligned}
\end{equation}
The idea on the derivation of \eqref{eq_se_grad_W} is described in Appendix \ref{partial_derivatives} to split the quartic expression into two quadratic expressions. For an initial estimate of the parameters, $[\textbf{W}_0, B_{h, 0}, B_{v, 0}]$, we use the following definitions for convenience
\begin{equation}
\label{eq_map_def2}
\begin{aligned}
& \widetilde{Y_{0}}=\textbf{W}_{0}{X}+{B_{h,0}} \textrm{ and } {Y_{0}}={A_{h}}(\widetilde{Y_{0}}) \\
& \widetilde{X_{0}}=\textbf{W}_{0}^{T}{Y_{0}}+{B_{v,0}} \textrm{ and } {X_{0}}={A_{v}}(\widetilde{X_{0}}) \\
& {\delta X}={A_{v}^{'}}(\widetilde{X_{0}})\odot ({X_{0}}-{X})
\end{aligned}
\end{equation}
With the definitions in \eqref{eq_map_def2} and a learning rate of $\gamma$, the GD learning for minimization \eqref{eq_se_min} can be described in the following steps to obtain a new update $[\textbf{W}_{1}, {B_{h,1}}, {B_{v,1}}]$:
\begin{equation}
\label{eq_se_gd}
\begin{aligned}
&{B_{v,1}}={B_{v,0}}-\gamma {\delta X} \\
&{B_{h,1}}={B_{h,0}}-\gamma {A_{h}^{'}}(\widetilde{Y_{0}})\odot [\textbf{W}_{0}{\delta X}] \\
&\textbf{W}_{1}=\textbf{W}_{0}-\gamma \{{{Y_{0}}{\delta X}^{T}+[{A_{h}^{'}}(\widetilde{Y_{0}})\odot (\textbf{W}_{0}{\delta X})]{X}^{T}}\} \\
\end{aligned}
\end{equation}
We point out that the GD for minimizing \eqref{eq_se_modified} is just replacing $({X_{0}}-{X})$ by $({X_{0}}-{A_{v}}({X}))$ in \eqref{eq_map_def2}. Recall ${Y_{0}}={A_{h}}(\textbf{W}_{0}{X}+{B_{h,0}})$ of \eqref{eq_map_def2}, we introduce
\begin{equation}
\label{eq_map_def3}
{Y_{1}}={A_{h}}(\textbf{W}_{0}{X_{0}}+{B_{h,0}})
\end{equation}
Here, we use the same notation to compare the CD-0 learning with the GD learning \eqref{eq_se_gd}. Let $\eta $ be the learning rate, the CD-0 for a singe sample can be expressed as
\begin{equation}
\label{eq_se_cd}
\begin{aligned}
&{B_{v,1}}={B_{v,0}}-\eta ({X_{0}}-{X}) \\
&{B_{h,1}}={B_{h,0}}-\eta ({Y_{1}}-{Y_{0}}) \\
&\textbf{W}_{1}=\textbf{W}_{0}-\eta ({Y_{1}}{X_{0}^{T}}-{Y_{0}}{X^{T}})
\end{aligned}
\end{equation}
When estimated $X_0$ is near the original $X$, $||X_{0}-X||$ may be small so that the second-order term $||\textbf{W}_0({X_{0}}-{X})||^2$ can be ignored numerically. Without the loss of generality, we assume ${A_{v}}(\cdot )$ is identity, then ${\delta X} = (X_0-X)$ and the first equation in \eqref{eq_se_gd} and \eqref{eq_se_cd} is equivalent. Next we compare the expressions of $B_{h,1}$ and $\textbf{W}_{1}$. By using the Taylor series expansion, we obtain
\begin{equation}
\label{eq_diff_Y}
\begin{aligned}
{Y_{1}}-{Y_{0}}&={A_{h}}(\textbf{W}_{0}{X_{0}}+{B_{h,0}})-{A_{h}}(\textbf{W}_{0}{X}+{B_{h,0}}) \\
&\approx {A_{h}^{'}}(\textbf{W}_{0}{X}+{B_{h,0}})\odot [\textbf{W}_{0}({X_{0}}-{X})] \\
&= {A_{h}^{'}}(\widetilde{Y_{0}})\odot [\textbf{W}_{0}({X_{0}}-{X})]
\end{aligned}
\end{equation}
\begin{equation}
\label{eq_diff_XY}
\begin{aligned}
{Y_1}{X_0^T}-{Y_0}{X^T}&={Y_0}{X_0^T}-{Y_0}{X^T} + {Y_1}{X_0^T}-{Y_0}{X_0^T} \\
&={Y_0}(X_0-X)^T+({Y_1}-{Y_0}){{X}}^T+({Y_1}-{Y_0})({X_0-X})^T \\
&\approx {Y_0}({X_{0}}-{X})^{T}+\{{{A_{h}^{'}}(\widetilde{Y_0})\odot [\textbf{W}_0({X_{0}}-{X})]}\}{X}^{T} 
\end{aligned}
\end{equation}

\noindent
Relationships \eqref{eq_diff_Y} and \eqref{eq_diff_XY} indicate that the CD-0 defined by \eqref{eq_se_cd} is an approximation of the GD learning of \eqref{eq_se_gd} by ignoring the higher order terms in the Taylor series expansion. For general activation ${A_{v}}(\cdot )$, the above argument holds through replacing $({X_{0}}-{X})$ by ${A_{v}^{'}}(\cdot )\odot({X_{0}}-{X})$. 

\noindent
{\bf Remark 3.} \emph{ Even in the linear case, the cost function (\ref{eq_se}) is quartic with respect to weight matrix ${\bf W}$, and is more complicated than the quadratic cost function in the feedforward neural networks after applying the nonlinear activation on both layers. For the first time, an explicit formula for the gradient of (\ref{eq_se_grad_W}) was derived in this paper but it is unknown whether the cost function (\ref{eq_se}) is concave to guarantee the convergence of the GD iterations. More discussions on the gradient descent method to find global and local minimums can be found in Section 4.3 of [\cite{Goodfellow15}]. Previously we proved that CD-0 is convergent for the conditional likelihood due to the convergence of EM algorithm, and is an approximate learning procedure of GD without considering the second-order errors. To achieve a global convergence numerically, we alternate the GD and CD iterations in the training experiments in next section.}

It is worth noting that the CD-0 defined in \eqref{eq_se_cd} can be extended to the directed network as well. Assume that $\textbf{W}$ is the weight matrix and ${Y}$ is the target, we consider the following squared error
\begin{equation}
\label{eq_se_slp}
E(X;\, \textbf{W}, {B})=\frac{1}{2}\Vert 
{A}(\textbf{W}{X}+{B})-{Y}\Vert ^{2}
\end{equation}
Here, ${B}$ is the bias and ${A}$ stands for the activation function and \eqref{eq_se_slp} is the squared error. For an initial estimate \{${X_{0}}$, ${B_{0}}$, $\textbf{W}_{0}$\} and ${Y_{1}}={A}(\textbf{W}_{0}{X_{0}}+{B_{0}})$, the CD-0 for the feedforward network can be described in the following steps:
\begin{equation}
\label{eq_se_slp_cd}
\begin{aligned}
&{X_{1}}={X_{0}}-\eta \textbf{W}_{0}^{T}({Y_{1}}-{Y}) \\
&{B_{1}}={B_{0}}-\eta ({Y_{1}}-{Y}) \\
&\textbf{W}_{1}=\textbf{W}_{0}-\eta ({Y_{1}}-{Y}){X_{0}^{T}} 
\end{aligned}
\end{equation}

Under the Gibbs distribution in section \ref{probabilistic_rbm}, we reformulate the general likelihood function (\ref{eq_likelihood_eqn}) into a conditional log-likelihood function (\ref{eq_loglikeli_cond}). In order to extend RBM for the data following other distributions from continuous variables, more analysis and modification of the likelihood functions can be found in [\cite{Woodford06, Hinton10, Fischer14}]. Here we provide a statistic interpretation of (\ref{eq_se}) by a posterior probability for Gaussian variables. Assume that $X\in S$ is one observation of the Gaussian variable and the reconstruction by \eqref{eq_mapping_yx} is an estimate of the mean, the probability of such observation is expressed as
\begin{equation}
\label{eq_post_prob}
\begin{aligned}
p(X;\, \Theta) =\frac{1}{\sqrt{(2\pi)^m}} \exp\left[-\frac{1}{2}\Vert 
{A_{v}}(\textbf{W}^{T}{A_{h}}(\textbf{W}{X}+{B_{h}})+{B_{v}})-{X}\Vert ^{2}\right]
\end{aligned}
\end{equation}
Please keep in mind that $Y$ is regarded as an intermittent parameter for estimating the mean of Gaussian variables. The likelihood function over all training data $S$ becomes
\begin{equation}
\label{eq_likelihood_gaus}
 L(\boldsymbol{\Theta};\, X) = {\mathop{\mathrm{\prod }}_{X\in S} p(X;\, \Theta)}
\end{equation}
With the concept of a posterior probability, the likelihood function (\ref{eq_post_prob}) becomes much simpler than the treatment in [\cite{Woodford06, Hinton10, Fischer14}] for continuous variables. This also can be regarded as an alternate approach on the conditional RBM investigated in [\cite{Mnih11}].

\section{Training Experiments for Scalar and Vector Nodes} \label{train_config}
We use several datasets including the well-known MNIST and CIFAR-10 benchmark data to demonstrate how to train the reformulated RBM with scalar and vector nodes. To leverage the global convergence of CD-0 algorithm for conditional likelihood and the faster convergence from GD, we alternate CD-0 and GD in our experiments. In subsection 4.3, we apply the Karhunen–Loève theorem to define the dimension of hidden layer. The configuration of RBM in subsection 4.2 provides a unique network architecture for correlation analysis.

\subsection{Scalar Variables} \label{image_analysis}
\label{unified_framework}
Dimensionality reduction has been one of most studied research topics in the field of machine learning and statistics, which can extract the key information from data [\cite{Akaike76, Hérault84, Müller05, Hinton06, Maaten08, Bu15, Jolliffe16, McInnes18, Edelmann19, Maaten09, Zebari22}]. Among the rich library, principal component analysis (PCA) has been the most used method and continues to be explored due to its simplicity and interpretability.  While some numerical experiments showed that the stacked RBMs can outperform PCA [\cite{Hinton06, Salakhutdinov09}], there are some investigations indicating that the performance of RBM over PCA is data dependent [\cite{Bu15}]. In this section, we compare PCA with linear and nonlinear RBM model by using GD iteration \eqref{eq_se_gd} and CD-0 iteration \eqref{eq_se_cd} in an alternating fashion, where the linear RBM uses the identity as the activation function and the nonlinear ones use other activation functions, such as sigmoid, softplus and relu. MNIST dataset (\url{http://yann.lecun.com/exdb/mnist/}) is used in the numerical experiments. All MNIST images have a fixed size of $28\times 28$ with integer values in $[0, 255]$, and are scaled to be in $[0, 255/268]$ in all experiments. The reason that we choose $255/268 \approx 0.95$ is to avoid the saturation of the activation functions in order to reach maximum $1.0$ for values going to the infinite. In summary, we found that the linear RBM achieves similar performance with PCA, while the nonlinear RBM with different activation functions lead to different performance. This demonstrates the reformulated RBM can select the ideal activation functions, depending on the data, to improve the performance in dimensionality reduction, which cannot be achieved by the original probabilistic RBM model.

\paragraph{Linear dimensionality reduction}
When activation functions ${A_{h}}(\cdot )$ and ${A_{v}}(\cdot )$ are the identity, the data mappings by \eqref{eq_mapping_yx} are linear transformations, and the reconstruction error \eqref{eq_se} becomes
\begin{equation}
\label{eq_se_linear}
E(X;\, \textbf{W}, {B_{h}}, {B_{v}})=\frac{1}{2}\Vert 
\textbf{W}^{T}\textbf{W}{X}+\textbf{W}^{T}{B_{h}}+{B_{v}}-{X}\Vert ^{2}
\end{equation}
It is well-known that PCA is based on the covariance matrix and its eigen decomposition, and its solution can be described as the solution of the following constrained minimization problem [see {\it Section 2.12} in \cite{Goodfellow15}]
\begin{equation}
\label{eq_se_min_pca}
\mathop{\mathrm{argmin}}_{\textbf{W}, {B_{h}}, {B_{v}}}\mathop{\mathrm{\sum_{X\in S}}}E(X;\, \textbf{W}, {B_{h}}, {B_{v}}) \textrm{ subject to } \textbf{W}^T\textbf{W}=\textbf{I}
\end{equation}
This implies that PCA is the linear RBM with an orthogonality constraint on \textbf{W}, so the linear RBM should result in better or equivalent performance comparing to PCA in minimizing the reconstruction error. More discussions are summarized in Remark 3 to compare with other existing works [\cite{Greenacre83, Timothy06, Pueyo16, Renbin18}].

\begin{figure}[ht]
\hfill
\begin{center}
\centerline{\includegraphics[width=\columnwidth]{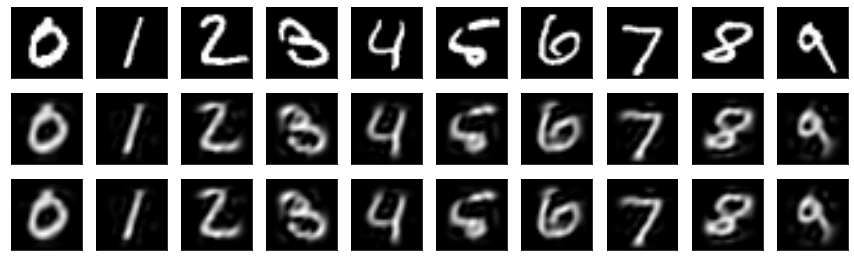}}
\vskip -0.05in
\caption{Original digits (top), reconstructed digits by linear RBM (middle), and reconstructed digits by PCA (bottom).}
\label{fg_lRBM1}
\end{center}
\end{figure}

In this subsection, we compare the linear RBM and PCA on MNIST dataset in two tasks: data reconstruction error and 2D data visualization capability. For data reconstruction, we set the dimension of the output as $49=7 \times 7$ which is relatively small for a single RBM compared with the stacked RBMs [\cite{Hinton06, Salakhutdinov09}]. The results are illustrated in Figure \ref{fg_lRBM1}, while the percentages of mean squared errors (MSEs) over the original digits for linear RBM and PCA are $10.2656\%$ and $10.2606\%$, respectively, indicating comparable performance between linear RBM and PCA on this dataset.

\begin{figure}[ht]
\begin{center}
\centerline{\includegraphics[scale=0.35]{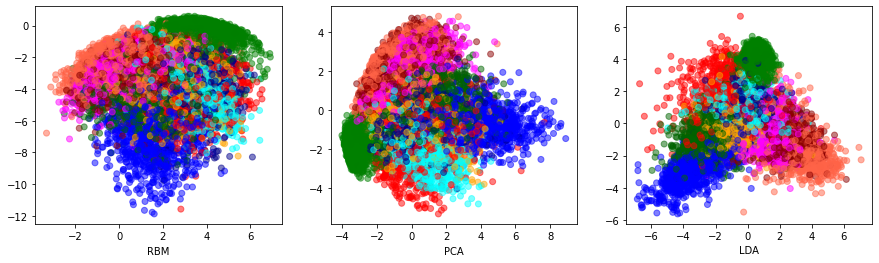}}
\vskip -0.05in
\caption{Distribution of digits by RBM (left), PCA (middle), LDA (right), while each color represents one digit.}
\label{fg_lRBM2}
\end{center}
\vskip -0.05in
\end{figure}

\noindent
For 2D data visualization, we set the dimension of hidden layer to be two and then visualize the dimension-reduced MNIST digits in the 2D plane to compare the digit distribution from linear RBM, PCA and linear determinant analysis (LDA) [\cite{Martinez01, McLachlan04}] in Figure \ref{fg_lRBM2}. We thus conclude that the linear RBM is comparable to PCA on the tasks of data reconstruction and 2D data visualization.

\begin{figure}[ht]
\vskip 0.1in
\begin{center}
\centerline{\includegraphics[scale=0.45]{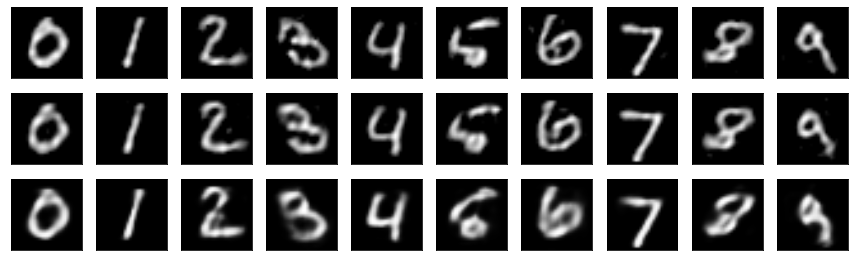}}
\caption{Reconstructed digits with softsign-relu (top), sigmoid-relu (middle), and sigmoid-sigmoid (bottom).}
\label{fg_nlRBM1}
\end{center}
\end{figure}

\noindent
{\bf Remark 4.} \emph{The non-negative matrix factorization may be translated into the following minimization problem with non-negative matrix.}
\begin{equation}
\mathop{\mathrm{argmin}}_{\textbf{W}, {B_{h}}, {B_{v}}}\mathop{\mathrm{\sum_{X\in S}}}E(X;\, \textbf{W}, {B_{h}}, {B_{v}}) \textrm{ subject to } \textbf{W} \ge 0
\label{eq_se_min_nmf}
\end{equation}
\emph{We point out that the multiplicative update rule in Lee and Seung [\cite{Lee00}] is similar to the multiplication based EM algorithm in the field of medical imaging [\cite{Shepp82, Hart87, You07}] for the data following the Poisson distribution. In addition to using the linear RBM with $L^2$ error to understand the traditional PCA, the linear RBM with $L^1$ error can be used to study the robust PCA [\cite{Candes09, Bouwmans18}] by solving the following minimization problem}
\begin{equation}
\mathop{\mathrm{argmin}}_{\textbf{W}, {B_{h}}, {B_{v}} } \mathop{\mathrm{\sum_{X\in S}}}\Vert 
\textbf{W}^{T}\textbf{W}{X}+\textbf{W}^{T}{B_{h}}+{B_{v}}-{X}\Vert_{L^1}
\label{eq_se_min_rpca}
\end{equation}
\paragraph{Nonlinear dimensionality reduction}
The nonlinear dimensionality reduction has been studied in the literature by using various disciplines [\cite{Scholkopf98, Hoffmann07, Gorban07, Maaten09, Scholz08, McInnes18, YShen18, XiaoZh2021, Nguyen21}]. In particular, several nonlinear representations, such as kernel PCA, t-distributed stochastic neighbor embedding (tSNE), and uniform manifold approximation and projection (UMAP) have been widely used and studied for visualization task [\cite{Scholz08, Hoffmann07, Maaten08, McInnes18}]. Here, we show that, with flexible selection of the activation functions in the reformulated RBM, improved nonlinear representations can be achieved in addition to these well-known algorithms while nonlinear RBM is not suitable for visualization.
\begin{table}[ht]
\vskip 0.10in
\renewcommand{\arraystretch}{1.0}
\begin{center}
\begin{tabular}{|c|c|c|}\hline
Model & Activations & MSE Percent \\ \hline
RBM(784, 49) & softplus-relu & $7.9956\%$ \\ \hline
RBM(784, 49) & sigmoid-relu & $8.7749\%$ \\ \hline
RBM(784, 49) & sigmoid-sigmoid & $19.7171\%$ \\ \hline
PCA & components=49 & $10.2606\%$ \\ \hline
\end{tabular}
\end{center}
\caption{MSE percentages from three pairs of activation functions for MNIST dataset.}
\label{algo_comp_mse}
\end{table}
Let $RBM(m,n)$ stand for one RBM with dimensions of $m$ and $n$ on visible and hidden layers, respectively. The deep Boltzmann machine (DBM) defined in [\cite{Hinton06, Salakhutdinov09}], also known as the stacked RBMs, is denoted by $DBM(...)$, for example, $DBM(784, 784, 196, 49)$ stands for three connected RBMs: $RBM(784, 784)$, $RBM(784, 196)$, and $RBM(196, 49)$. In the numerical experiments for RBM, we have presented the results from three pairs of activation functions as listed in Table \ref{algo_comp_mse}. The reconstructed digits are shown in Figure \ref{fg_nlRBM1}. The first two pairs, i.e., softplus-relu and sigmoid-relu, are better than PCA while the third pair is much worse. This demonstrates the capability of reformulated RBM in selecting different activation functions to achieve a better performance.
\begin{figure}[ht]
\vskip -0.1in
\begin{center}
\centerline{\includegraphics[scale=0.4]{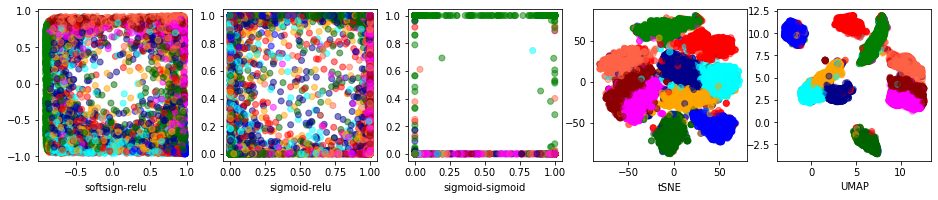}}
\caption{Spatial distribution of digits in 2D plane by RBM (softsign-relu), RBM (sigmoid-relu), RBM (sigmoid-sigmoid), tSNE and UMAP from left to righ.}
\label{fg_nlRBM2}
\end{center}
\end{figure}
For the visualization by projecting data to the 2D plane, we compare the nonlinear RBM with three pairs of activation functions over tSNE and UMAP in Figure \ref{fg_nlRBM2}. Three different pairs of activation functions produced very different spatial distributions, which implies the importance of selecting the suitable activation functions toward a good performance in dimensionality reduction. We conclude that the nonlinear RBM is designed for reducing the reconstruction error and thus is not suitable for 2D data visualization, which was the designed purpose of nonlinear methods tSNE and UMAP.

\subsection{Vector Variables} \label{multivariate_analysis}
Till this end, only the scalar value is considered as the input of each node in the RBM model. In this section, we extend the data mapping to vectors through tensor product. Let $\otimes$ denote the direct product between two vector spaces. For the RBM with vector-valued nodes, ${\bf X}$ and ${\bf Y}$ can be expressed by the direct product of two vector spaces, indicating ${\bf X}={X_1}\otimes {X_2} \in R^{m_1}\otimes R^{m_2}$ and ${\bf Y}={Y_1}\otimes {Y_2} \in R^{n_1}\otimes R^{n_2}$. Accordingly, the weight matrix ${\bf W}$ can be decomposed into two separate matrices ${\bf W}_1\in R^{n_1}\times R^{m_1}$ and ${\bf W}_2\in R^{n_2}\times R^{m_2}$ for separate linear transformations between $R^{m_1}$ and $R^{n_1}$ and between $R^{m_2}$ and $R^{n_2}$, respectively. The data mappings in \eqref{eq_mapping_yx} for vector-valued nodes are expressed as
\begin{equation}
\label{eq_map_vec_y}
{\bf Y}={A_h}([{\bf W}_1\times {X_1}]\otimes [{\bf W}_2\times {X_2}]+B_{1, h}\otimes B_{2, h})
\end{equation}
\begin{equation}
\label{eq_map_vec_x}
{\bf X}={A_v}([{\bf W}_1^T\times {Y_1}]\otimes [{\bf W}_2^T\times {Y_2}]+B_{1, v}\otimes B_{2, v}) 
\end{equation}

\paragraph{Linear correlation analysis}
Now we configure the mappings \eqref{eq_map_vec_y} and \eqref{eq_map_vec_x} to find the number of the independent sequences, which is one type of independent component analysis [\cite{Hérault84, Zebari22}]. We let ${\bf W}_2={\bf I}$, and each element of ${X}$ and ${Y}$ will become a vector with the same dimension, and equations \eqref{eq_map_vec_y} and \eqref{eq_map_vec_x} can be rewritten as
\begin{equation}
\label{eq_vec_corr_yx}
{Y^{j}}=\sum_{i}{w_{j, i}{X^{i}}+{B_{h}^{j}}}, \quad {X^{i}}=\sum_{j}{w_{j, i}{Y^{j}}+{B_{v}^{i}}}
\end{equation}

\begin{table}[ht]
\vskip -0.10in
\renewcommand{\arraystretch}{0.8}
\begin{center}
\begin{tabular}{|l|l|l|} \hline
Name & Distribution & Data Generation Method\\ \hline
S01 & Poisson & poisson(3.0, 2000) \\ \hline
S02 & Binomial & binomial(10.0, 0.6, 2000) \\ \hline
S03 & Laplace & laplace(-1.0, 1.0, 2000) \\ \hline
S04 & Normal & normal(0.5, 1.0, 2000) \\ \hline
S05 & Exponential & exponential(2.0, 2000) \\ \hline
S06 & Uniform & uniform(-2.0, 2.0, 2000) \\ \hline
S07 & Mixture & S01$\times$0.25 + S02$\times$0.75 + S03$\times$0.50 \\ \hline
S08 & Mixture & S02$\times$0.30 + S03$\times$0.70 + S04$\times$0.50 \\ \hline
S09 & Mixture & S03$\times$0.45 + S04$\times$0.55 + S05$\times$0.35 \\ \hline
S10 & Mixture & S04$\times$0.60 + S05$\times$0.40 + S06$\times$0.20 \\ \hline
S11 & Mixture & S05$\times$0.50 + S06$\times$0.35 + S01$\times$0.45 \\ \hline
S12 & Mixture & S06$\times$0.40 + S01$\times$0.10 + S02$\times$0.60 \\ \hline
\end{tabular}
\vskip 0.10in
\caption{random sequences and combinations.}
\label{seq_rnd}
\end{center}
\vskip -0.10in
\end{table}
\noindent
For the numerical experiment, we use python numpy package to generate six independent random sequences with 2000 points following the Poisson, binomial, Laplace, Gaussian, exponential, and uniform distributions and then make another six sequences by combining them with linear operations (Table \ref{seq_rnd}). Since the ground truth is that there are six independent sequences, if the number of nodes in $Y$ is less than six, the reconstruction can not be accurate and the error will be significant. We use a window length of 50 and the stride length of 20 to generate training subsequences through moving the window across the 2000 points. In this configuration, $X$ has 12 nodes with a 50-dimensional vector on each node, $Y$ will have various numbers of nodes with a 50-dimensional vector on each node.
\begin{figure}[ht]
\centerline{\includegraphics[scale=1.0]{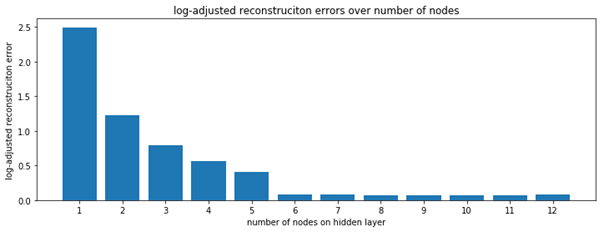}}
\vskip -0.05in
\caption{log-adjusted reconstruction error.}
\label{fg_log_error_seq}
\end{figure}
As shown in Figure \ref{fg_log_error_seq}, using the log-adjusted error $-1.0/\log(MSE)$ as the reconstruction errors, we found that the configuration of \eqref{eq_vec_corr_yx} can be used to find the number of independent sequences and the coefficients between the independent and dependent sequences for linear correlations. We point out that the nonlinear combinations of independent sequences can be studied with non-identity activation functions.

\begin{figure}[ht]
\begin{center}
\centerline{\includegraphics[scale=0.45]{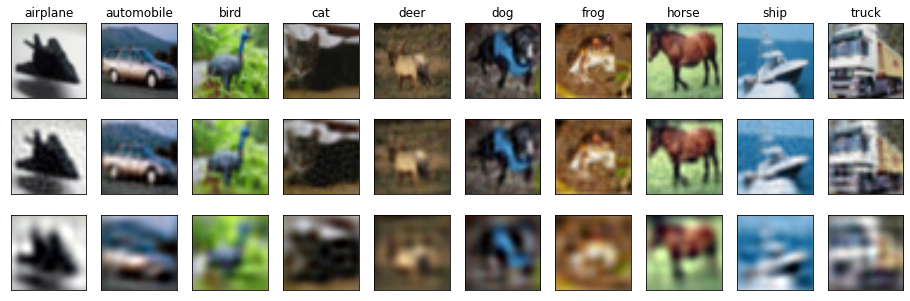}}
\vskip -0.05in
\caption{Original images (top), reconstructed images for $256$ (middle) and $64$ (bottom) nodes in the hidden layer, respectively.}
\label{fg_lRBM3}
\end{center}
\vskip -0.25in
\end{figure}

\paragraph{Dimensionality reduction for color images}
Next, we perform the dimensionality reduction for color images with the reformulated RBM. While each pixel in the gray scale images is a scalar, the pixels of color images are three-dimensional vectors, which cannot be processed by the traditional RBM model directly. We let ${\bf W}_1={\bf I}$, and each element of ${X}$ and ${Y}$ is a vector with different dimensions but the number of nodes on visible and hidden layers remains to be 3. Using the block matrix operation, equations \eqref{eq_map_vec_y} and \eqref{eq_map_vec_x} can be rewritten as:
\begin{equation}
\label{eq_vec_feat_yx}
Y^i = \textbf{W}X^i + B_h^i, \quad X^i = \textbf{W}^T Y^i + B_v^i
\end{equation}
For the numerical experiments, we use CIFAR-10 dataset, which includes total 60000 $32\times 32$ color images in 10 classes. The number of nodes on visible and hidden layers to be three, the dimension of vectors on visible layer is $964$, and the dimensions of vectors on the hidden layer will be $256$ and $64$ in two numerical experiments. The reconstructed images for the first appeared image of each class from test images of CIFAR-10 are shown in Figure \ref{fg_lRBM3}, which demonstrates the capability of the reformulated RBM to conduct dimensionality reduction for the color images.

\subsection{Selection of Reduction Dimension}
In the application of data dimensionality reduction, the dimension of original data is known but the target dimension to be reduced to is selected in an ad hoc way, depending on the nature of data and the experience of practitioners. In this paper, based on the Karhunen–Loève theorem [\cite{Stark86, Ghanem91}] we provide a theoretical framework to define the initial reduction dimension. After applying certain data centralization technique such as the z-score transform, we may assume that the average of $X$, $\overline{X} = 0$. Let $\{\lambda_k\}$ be the eigen values of the covariance matrix of $\{X_i\}$ in the decreasing order, and $\{e_k\}$ are the corresponding eigen vectors. Consider the following expansion
\begin{equation}
\label{eq_rank_decomp}
X_i = \sum_{k=1}^{M}c_{i, k}e_k + \epsilon_i
\end{equation}
Here $M\le m$, $m$ is the dimension of $X$. For a given ratio $\rho \in (0, 1)$, if we want $\sum{||\epsilon_i||^2_{L^2}} \le (1-\rho)\sum{||X_i||^2_{L^2}}$, the minimum $M(\rho)$ is defined by
\begin{equation}
\label{eq_rank_cutoff}
M(\rho) = \underset{M}{\operatorname{argmin}} \{\frac{\sum_{k=1}^{M}\lambda_k}{\sum_{k=1}^{m}\lambda_k} \ge \rho\}
\end{equation}
We call $M(\rho)$ as the approximate dimension of $\{X_i\}$ for a given error ratio of $(1-\rho)$ in the sense of total variance, and $M(\rho)$ will be used as the reduced dimension of the hidden layer. If the element of $X$ is a vector, the estimate remains valid after flattening $X$ to be a one-dimensional vector.

\section{Discussion}
In this work, we have revisited the probabilistic RBM model and provided the theoretical insights of the CD learning algorithm from the viewpoint of MAP and EM algorithm. With a conditional probability \eqref{eq_gibbs_cond}, CD-0 was proved to be convergent as one implementation of EM algorithm. We demonstrated that the reformulation of the RBM into a deterministic model offers several advantages, including the flexibility of the input data and the choice of the activation functions. Moreover, the reformulated RBM provides a unified architecture for dimensionality reduction. For instance, the PCA and robust PCA are actually the linear RBM, whose activation functions are identity, with different constraints. For data with nonlinear correlations, the nonlinear RBM can incorporate different activation functions to capture the nonlinearity and improve the performance in the dimensionality reduction although it is not suitable for 2D data visualization. Several interesting directions can be pursued based on the current work, including but not limited to: 1) using CD-0 \eqref{eq_se_slp_cd} to train the neural network models when the gradient is not available; 2) performing more numerical comparison studies to provide the advantages and disadvantages of the reformulated RBM over many existing works in the literature; 3) independent component analysis of the nonlinearly correlated data; and 4) investigating the numerical behavior of the RBM with the constraints and $L^1$ error defined in \eqref{eq_se_min_nmf} and \eqref{eq_se_min_rpca}.

The research on RBMs was very active in the early 2000s due to its improved dimensionality reduction in [\cite{Hinton06}] for pretraining deep neural networks and the application to recommendation systems in [\cite{Ruslan07}]. However, with the more advanced autoencoder networks such as [\cite{VincentLaro10, Welling19}], further studies on RBMs have decreased since 2010. Recently the Transformer from [\cite{Vaswani17, Choi20}] became the dominant architecture in the application of encoding and decoding vectorized text data, and the use of RBMs became much less active, we refer to [\cite{BondTaylor22, Zebari22}] for recent trends on the research of autoencoder. Thus, a substantial numerical comparison study was not performed in preparing this paper. Instead, the focus of this paper is to reformulate the probabilistic RBM by a conditional likelihood so that the convergence of CD can be studied through the EM algorithm, and to extend the RBM to vector variables with a deterministic model to avoid interpreting nodes by specific statistic distributions in the likelihood definition. Finally, we want to mention that the weights in the expression of direct products by (\ref{eq_map_vec_y}) and (\ref{eq_map_vec_x}) are a new extension of the classical RBM to handle continuous vector variables; in particular, the correlation analysis in subsection 4.2 could be a new application of the RBM for sequence data.
\appendix
\section{Gradients of matrix variables in an inner product}
\label{partial_derivatives}

For the linear case, we provide the steps on the calculation of gradients of \eqref{eq_se} with a complicated inner product between two vectors. Hereafter, we use $< , >$ for the inner product of two vectors for any dimension. Given vectors ${P}, {Q}\in R^m$, for a matrix variable $\textbf{W} \in R^{n\times m}$, we consider an inner product
\begin{align*}
I(\textbf{W}) &= <\textbf{W}{P}, \textbf{W}{Q}> \\
&= <{\textbf{W}}^T\textbf{W}{P}, {Q}>=<{\textbf{W}}^T\textbf{W}{Q}, {P}>
\end{align*}
The partial derivatives of $I(\textbf{W})$ with respect to $\textbf{W}$ are:
\begin{align*}
\frac{\partial I(\textbf{W})}{\partial \textbf{W}} = [\textbf{W}{P}]{{Q}}^T+[\textbf{W}{Q}]{{P}}^T
\end{align*}
\noindent
Notice that there are two terms on the right side, which explains why we have two components in equation \eqref{eq_se_grad_W}. In the linear case of ${B_{h}}=0$ and ${B_{v}}=0$, equation \eqref{eq_se} degenerates to
\begin{align*}
E(\textbf{W}) = \frac{1}{2}<\textbf{W}^{T}\textbf{W}{X}-{X}, \textbf{W}^{T}\textbf{W}{X}-{X}>
\end{align*}
Let ${\Delta X}=\textbf{W}^{T}\textbf{W}{X}-{X}$ and consider the symmetries of the inner product. We obtain:
\begin{align*}
\frac{\partial E(\textbf{W})}{\partial \textbf{W}} 
&= \frac{\partial }{\partial \textbf{W}}<\textbf{W}^{T}\textbf{W}{X}-{X}, {\Delta X}> \\
&=[\textbf{W}{X}]{{\Delta X}}^T+[\textbf{W}{\Delta X}]{{X}}^T
\end{align*}
If the activation functions are not the identity, the scaling factors introduced by the derivatives of activation functions must be applied to the matrix multiplications accordingly.

\bibliography{refs}

\end{document}